%% file: main.tex
\documentclass[letterpaper, 10 pt, conference]{ieeeconf}  % Comment this line out if you need a4paper

\IEEEoverridecommandlockouts                              % This command is only needed if 
\usepackage{amsmath,amsfonts}
\usepackage{algorithmic}
\usepackage{array}
\usepackage[caption=false,font=normalsize,labelfont=sf,textfont=sf]{subfig}
\usepackage{textcomp}
\usepackage{stfloats}
\usepackage{url}
\usepackage{verbatim}
\usepackage{graphicx}
\usepackage{cite}
\usepackage{tabularx}
\usepackage{amssymb}
\usepackage[linesnumbered,ruled]{algorithm2e}
\usepackage{multirow}
\usepackage{multicol}
\usepackage{diagbox}
\usepackage{makecell}
\usepackage{indentfirst}
\usepackage{bbding}
\usepackage{booktabs}
\usepackage{hyperref}
\usepackage{cleveref}

\title{\LARGE \bf
AdvDiffuser: Generating Adversarial Safety-Critical Driving Scenarios via Guided Diffusion}

\author{Yuting Xie$^{1}$, Xianda Guo$^{2}$, Cong Wang$^{3}$, Kunhua Liu$^{4}$ and Long Chen$^{3\dag}$% <-this % stops a space
\thanks{This work was supported by the National Natural Science Foundation of China (Grant No.62373356), and the Science \& Technology Support Project for Young People in Colleges of Shandong Province (Grant No. 2023KJ120).}% <-this % stops a space
\thanks{$\dag$ Corresponding author: L. Chen.}
\thanks{$^{1}$Y. Xie is with the School of Computer Science and Engineering, Sun Yat-Sen University
    {\tt\small xieyt8@mail2.sysu.edu.cn}}%
\thanks{$^{2}$X. Guo is with the School of Computer Science, Wuhan University
        {\tt\small xianda\_guo@163.com}}%
\thanks{$^{3}$C. Wang and L. Chen are with the State Key Laboratory for Management and Control of Complex Systems at the Institute of Automation, Chinese Academy of Sciences
        {\tt\small wangcong2024@ia.ac.cn}
        {\tt\small long.chen@ia.ac.cn}}%
\thanks{$^{4}$K. Liu is with the School of Mechanical and Automotive Engineering, Qingdao University of Technology
        {\tt\small liukh@qut.edu.cn}}%
}

\begin{document}

\maketitle
\thispagestyle{empty}
\pagestyle{empty}

\input{sec/0_abstract}    
\input{sec/1_intro}

\input{sec/2_related}
\input{sec/3_method}

\input{sec/4_exp}

\input{sec/5_dis}

\clearpage
{
    \small
    \bibliographystyle{IEEEtran}
    \bibliography{IEEEabrv, main}
}


\end{document}

%% file: sec/0_abstract.tex
\begin{abstract}
Safety-critical scenarios are infrequent in natural driving environments but hold significant importance for the training and testing of autonomous driving systems. The prevailing approach involves generating safety-critical scenarios automatically in simulation by introducing adversarial adjustments to natural environments. These adjustments are often tailored to specific tested systems, thereby disregarding their transferability across different systems.
In this paper, we propose AdvDiffuser, an adversarial framework for generating safety-critical driving scenarios through guided diffusion. By incorporating a diffusion model to capture plausible collective behaviors of background vehicles and a lightweight guide model to effectively handle adversarial scenarios, AdvDiffuser facilitates transferability. Experimental results on the nuScenes dataset demonstrate that AdvDiffuser, trained on offline driving logs, can be applied to various tested systems with minimal warm-up episode data and outperform other existing methods in terms of realism, diversity, and adversarial performance.
\end{abstract}

%% file: sec/1_intro.tex
\section{Introduction}
\label{sec:intro}
Safety evaluation of autonomous vehicles (AV) requires scalable long-tail driving scenarios~\cite{kang2019test, ding2023survey, chen2022milestones}. However, these kinds of scenarios are rare in the real world, which poses a data-rarity problem. A prevailing alternative is to generate safety-critical scenarios in simulation. Rather than manually designing scenarios from scratch~\cite{zhang2022rethinking}, recent works seek to autonomously generate challenging scenarios via perturbing existing scenarios~\cite{mixsim,strive,advsim,drl}. Generally, these works modify maneuvers of a single or a small group of background vehicles (BV) via adversarial reinforcement learning \cite{drl} or optimization searching on a scenario parameterization space ~\cite{mixsim,strive,advsim}. Since the specific tested AV system is in the loop, these studies lack investigation into the transferability across diverse types of target AVs, which leaves the generated scenarios less flexible.

Recently, diffusion models have made significant progress in vision and language tasks \cite{saharia2022photorealistic,rombach2022high}, showcasing their potential in few-shot or zero-shot learning as a highly promising generative model. And several studies have emerged that employ diffusion models to address sequence decision problems \cite{diffuser,RN228,RN214,RN224,metadiffusion}, wherein guidance from an auxiliary reward function is injected into the sampling process to produce class-conditional outcomes. These works demonstrate the powerful generative capabilities of diffusion models in handling out-of-distribution data without retraining.

\begin{figure}[t]
\centering
\includegraphics[width=1.0\linewidth]{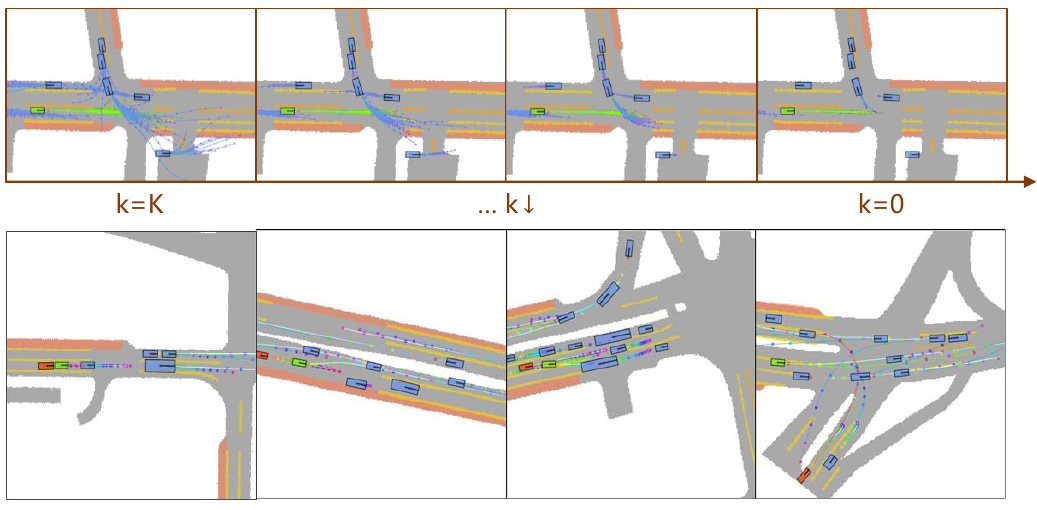}
   \caption{AdvDiffuser generates safety-critical scenarios for testing AV systems. (Top) Diffusion-based traffic simulation involves noise undergoing k rounds of reverse diffusion process. (Bottom) The diffusion model, coupled with a reward guide, generates adversarial background trajectories. AVs are depicted in green, while vehicles executing attack behaviors are in orange.}
\label{fig:illustration}
\vspace{-0.6cm}
\end{figure}

Inspired by this, we present a novel framework for generating safety-critical driving scenarios called AdvDiffuser, which utilizes a guided diffusion model to generate adversarial trajectories. AdvDiffuser targets the planner component of AV systems, which includes prediction, planning and control modules, by planning behaviors of BVs online to intentionally disrupt targeted planners and induce collisions.
As depicted in Fig~\ref{fig:illustration}, AdvDiffuser learns adversarial collective behaviors for BVs from offline driving logs and undergoes an interactive 'warm-up' process for online scenario generation, allowing it to adapt to various target planners effectively. 
The core idea of AdvDiffuser involves decoupling realism and adversariness within a diffusion model and an auxiliary collision reward model to generate plausible multi-agent trajectories while promoting the creation of adversarial trajectories.
The advantage of this decoupling lies in the fact that only a small scale of parameters in the reward model needs adjustment when adapting to new targets, enabling online adversarial training for continuous improvement of autonomy.

We summarize the main contributions of this paper as follows. (1) By incorporating guided sampling into the driving simulation, we present a novel framework for generating safety-critical scenarios. (2) A multi-vehicle traffic simulation method based on a diffusion model is proposed, which effectively generates diverse and realistic background traffic flows. (3) Experimental results confirm the transferability advantages of our approach, highlighting its practical benefits in autonomous driving testing.

%% file: sec/2_related.tex
\section{Related Work}
\label{sec:related-work}

\subsection{Testing Autonomous Driving in Simulation}
Conventional self-driving vehicle simulation testing relies on log replay, lacking reactivity and causing domain shift issues~\cite{li2019aads, kar2019meta,codevilla2019exploring}. Efforts to address this include integrating reactivity into driving logs for closed-loop safety evaluation~\cite{nde, drl, advsim, closed-loop}. Early heuristic simulators~\cite{carsim, carla, sumo} have been replaced by neural traffic models like SimNet~\cite{simnet} and TrafficGen~\cite{trafficgen}, which mimic human driving behaviors but lack controllability. Mixsim~\cite{mixsim} offers controllability with constrained diversity, while Strive~\cite{strive} employs a VAE-based network for plausible traffic modeling, albeit compromising realism. In contrast, our method models traffic flow as a diffusion process, using VAE's latent codes to balance realism and diversity, thus enabling controllability through an auxiliary reward function.

\subsection{Safety-critical Driving Scenario Generation}
Simulating failure scenarios is crucial for comprehensive risk assessment in autonomous systems, particularly due to autonomous systems' superior performance in controlled environments \cite{ding2023survey}. Manual creation of challenging scenarios faces scalability issues and may lead to unrealistic situations \cite{waymo, metadrive, closed-loop}. Recent studies focus on parameterization spaces to identify adversarial parameters using optimization-based methods \cite{advtraj,advsce,RN246,RN247, RN249,RN250,advsim}. Methods like AdvSim directly perturb trajectory space while maintaining physical feasibility \cite{advsim}, while others optimize parameters within a latent space \cite{RN246, mixsim, strive}. Nevertheless, these methods require iterative re-planning, leading to efficiency issues.

Adversarial policy models, such as NADE and D2RL, train vehicles to execute adversarial maneuvers, offering flexibility but introducing complexity and requiring a substantial number of interactions \cite{ppo, ade, drl}. Our approach utilizes lightweight RL to guide sampling towards safety-critical variations efficiently, leveraging a diffusion formulation for seamless transfer.

\subsection{Diffusion Models for Sequence Decision Problems}
Conditional diffusion models can typically be categorized into two types: classifier-guidance and classifier-free~\cite{yang2022diffusion}. The former improves sampling quality in a specific domain by utilizing gradients from a pre-trained classifier, while the latter directly incorporates class-related context into the noise model, eliminating the need for such a pre-trained classifier. These models excel in vision and language tasks and are now applied in sequence decision problems. Diffuser~\cite{diffuser} employs a diffusion model to strategize robot behaviors, which is trained through a diffusion process over random noisy trajectories. It follows the classifier-guidance manner by incorporating gradients from a separately pre-trained reward model. Gu et al.~\cite{RN228} integrate maps and nearby agent states, directly into the diffusion model, to tackle the complex task of predicting multi-pedestrian trajectories. 
Combining classifier-guidance and classifier-free, M-Diffuser develops a constrained sampling framework for controllable trajectory generation in multi-agent scenarios using learnable differentiable cost functions. Instead of learned cost functions, Zhong et al.~\cite{RN224} use analytical loss functions derived from Signal Timing Logic (STL) rules to control trajectories.

Similar to \cite{RN214, RN224}, our approach is built upon guided diffusion formulation, together with incorporating contextual information in a classifier-free manner. But unlike previous attempts that merely touched upon the controllability of trajectory sampling, we are among the first to incorporate conditional diffusion into safety-critical driving scenarios generation tasks. Furthermore, we investigate the generalization potential of diffusion models in this field, an aspect overlooked by most previous works. In this context, our work bears the closest resemblance to MetaDiffusion~\cite{metadiffusion}, which leverages a conditioned diffusion model to plan robot behaviors, facilitating generalization across tasks that were previously unseen. Differently, our work focuses on the challenge of generalizing across unknown adversarial targets within an adversarial task setting.

%% file: sec/3_method.tex
\begin{figure*}[htbp]
\centering
\includegraphics[width=0.92\linewidth]{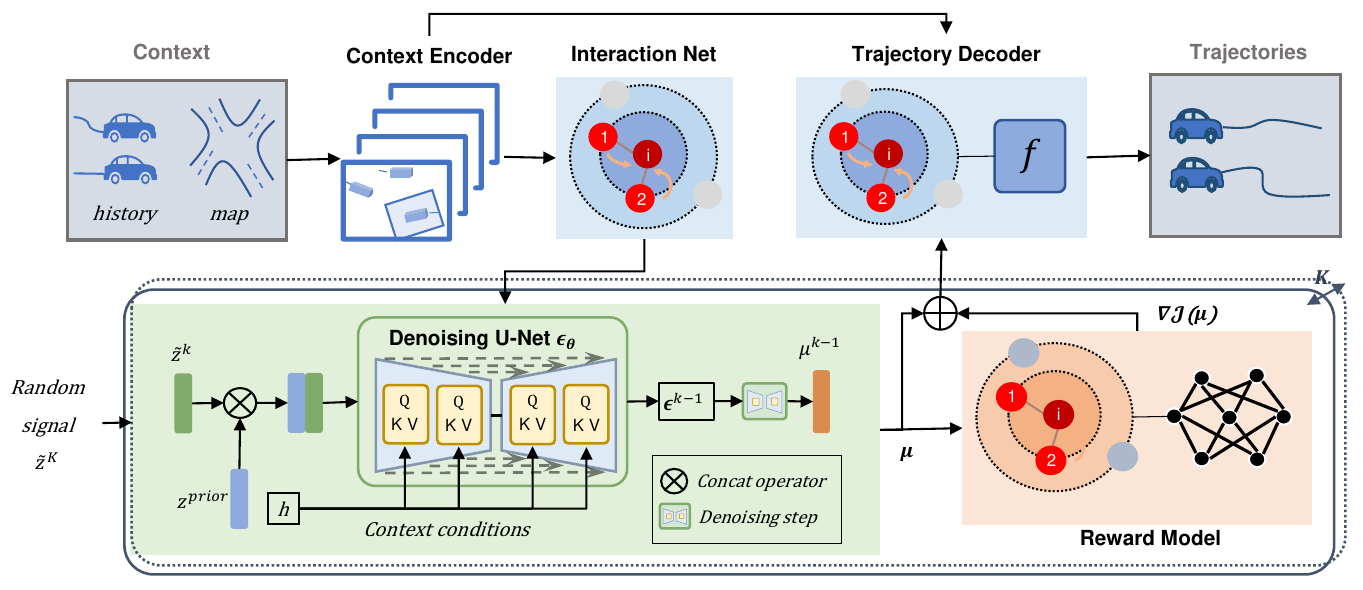}
   \caption{Overview of our AdvDiffuser framework. Given a context comprising historical vehicle trajectories and the map, AdvDiffuser aims to generate adversarial trajectories for background vehicles. The Interaction Net is a Graph Neural Network (GNN) architecture that encodes inter-vehicle interactions, providing a latent code $z_{prior}$ as a condition for the denoising process. The dynamics model $f$ in Trajectory Decoder transforms vehicle actions into trajectories.}
\label{fig:pipeline}
\vspace{-0.5cm}
\end{figure*}

\section{AdvDiffuser}
\subsection{Problem Formulation}
% m past 上标k，下标t,i.轨迹. 目标是找到最佳背景轨迹使目标车辆表现最差，撞墙或者与其他车辆发生碰撞。
Our goal is to create realistic challenging testing scenarios that induce failures in tested autonomous driving systems.
In particular, a driving scenario $\mathcal{S}$ consists of a high-definition map $M$ containing semantic information for drivable areas and lanes, agent states $x_{0:T}$(or trajectory ${\boldsymbol{\tau}}$), agent actions $u_{0:T}$. We denote $x_t$ (resp., $u_t$) the joint states of all agents at timestep t. 
Similarly, we denote $m_t$ as the collection of all agents's surrounding maps and $Past$ as historical states. Further, we introduce a subscript $i$ to indicate the ith vehicle and employ a superscript `+` for tested autonomous vehicles to distinct them from background vehicles.
% Specifically, we parameterize agent state $x_{t,i}$ by its position $[x_{t,i}, y_{t,i}]$, heading $\theta_{t,i}$ and velocity $v_{t,i}$, while action $u_{t,i}$ by steering angle and acceleration. 
Specifically, we parameterize agent states by their 2D position, heading, and velocity, while actions by steering angle and acceleration. 

The objective of autonomous driving systems is to optimize actions through a cost function $\mathcal{C}$ to ensure comfortable and safe maneuvering. In contrast, our objective is to intentionally disrupt their comfort and security, maximizing this cost. As our method generates trajectories directly, which can be converted to actions through a dynamic model, we describe this objective using trajectory representation.
\begin{equation}
    {\boldsymbol{\tau}}^* = \mathop{max}\limits_{{\boldsymbol{\tau}}} C({\boldsymbol{\tau}}^+,{\boldsymbol{\tau}},M,Past)
\end{equation}

\subsection{Diffusion Model for Multi-agent Trajectories}
\label{subsec:diff-m}
We employ a diffusion model to generate plausible collective behaviors of background vehicles. The diffusion model conceptualizes data generation as an iterative denoising process, which is the inverse of a forward diffusion process with state transitions following the properties of a Markov chain. As the number of iterations increases, the model eventually converges to a stationary distribution. Therefore, by introducing a learnable noise prediction model, we can restore random signals within specified domains through the iterative denoising process. The diffusion process is implemented within the latent space in our multi-agent trajectory generation model, rather than directly on vehicle trajectories. To ensure the physical plausibility of the generated trajectories, we further integrate it with an additional autoregressive trajectory decoder.
% \begin{figure}[t]
% \centering
% \includegraphics[width=\linewidth]{figs/architecture-v1.2.pdf}
%    \caption{Network architecture. 
%    The denoiser module is built upon a U-Net architecture conditioned on context embeddings.  Within the modules of AdvDiffuser, information fusion among vehicles is realized through an interaction net, a graph neural network on a fully connected scene graph covering all agents.}
% \label{fig:network}
% \end{figure}
% \begin{figure}[htbp]
%   \centering
%   \vspace{-0.62cm}
%   \subfloat[Denoiser]
%   {\includegraphics[width=0.8\linewidth]{figs/arch-a.pdf}}
%   \vspace{.05in}  
%   \subfloat[Context encoder]
%   {\includegraphics[width=0.40\linewidth]{figs/arch-b.pdf}} 
%   \hspace{.15in}
%   \subfloat[Reward model]
%   {\includegraphics[width=0.40\linewidth]{figs/arch-c.pdf}}   
%   \caption{Network architecture. The denoiser is built upon a U-Net architecture conditioned on context embeddings. In both the context encoder and reward model, information fusion among vehicles is achieved through an interaction net, which is a graph neural network on a fully connected scene graph that covers all agents.}
%   \label{fig:network}
% \vspace{-1cm}
% \end{figure}

\textbf{Architecture.}
The process of sampling multiple vehicle trajectories involves a context encoder, a trajectory decoder, and a reverse diffusion process. 
The context encoder $E_\theta(m, past)$ takes surrounding maps and historical trajectories as input, generating trajectory context $(h, \bf{z})$. As illustrated Fig.~\ref{fig:pipeline}, a simple Multi-Layer Perceptron (MLP) encodes vehicle state embedding $h_i$, and an interaction net with two fully connected graph layers aggregates vehicle states, capturing interactions and producing latent codes $\bf z_i$ for each vehicle.

The reverse diffusion process models a conditional probability distribution $p_\theta(\tilde{\bf z}_i^{k-1}|\tilde{\bf z}_i^k,\bf{z}_i,h_i)$ to align with the posterior distribution $q(\tilde{\bf z}_i^k|{\boldsymbol{\tau}},h)$. Following \cite{ddim}, we formulate the denoising process with a noise function ${\boldsymbol{\epsilon}}^{k}$ as:
\begin{subequations}
    \begin{gather}
        \begin{split}
            {\boldsymbol{\mu}}^{k-1} 
            &= \sqrt{\alpha^{k-1}}\left({\frac{\bf \tilde{z}_i^{k}-\sqrt{1-\alpha^{k}}{\boldsymbol{\epsilon}}^{k}}{\sqrt{\alpha^{k}}}}\right)\\
            &+\sqrt{1-\alpha^{k-1}}{\boldsymbol{\epsilon}}^{k}
        \end{split}\\
        q({\bf \tilde{z}_i}^{k-1}|{\bf \tilde{z}_i}^{k};{\boldsymbol{\epsilon}}^{k}) = \mathcal{N}({\boldsymbol{\mu}}^{k-1}, \frac{1-\alpha^{k-1}}{1-\alpha^{k}}\beta^{k}) 
        \label{equ:transition}
    \end{gather}
\end{subequations}
where $k$ denotes the $kth$ round of diffusion, $\alpha^{k}=\prod_{s=1}^{k}{(1-\beta^{s})}$ and $\beta^{k}$ are fixed variance schedulers that controls scale of noise. By parameterizing the noise function through a trainable noise prediction network $\varepsilon_\theta$, we obtained the conditional distribution as: 
\begin{equation}
    p_\theta(\tilde{\bf z}_i^{k-1}|\tilde{\bf z}_i^k,{\bf z}_i,h_i)=
    q({\tilde{\bf z}_i}^{k-1}|{\tilde{\bf z}_i}^{k}; \varepsilon_\theta(\tilde{\bf z}_i^k,k,{\bf z}_i,h_i))
    % \left\{
    % \begin{aligned}
    %     \mathcal{N}(\varepsilon_\theta(\tilde{z}_i^1,1,z_i,h_i),0) \quad if \ t=1,\\
    %     q({\bf \tilde{z}_i}^{k-1}|{\bf \tilde{z}_i}^{k}; \varepsilon_\theta(\tilde{z}_i^k,k,z_i,h_i)) \quad otherwise\\        
    % \end{aligned}
    % \right
    .
\end{equation}
A U-Net architecture is employed in the noise prediction network as illustrated in Fig.~\ref{fig:pipeline}. 
To incorporate contextual information, ${\bf z}_i$ is concatrated with ${\bf \tilde{\bf z}_i}^k$ and integrated into $\varepsilon_\theta$. Additionally, an adaptive group normalization (AdaGN) structure~\cite{dhariwal2021diffusion} is introduced to the U-Net architecture for taking $h_i$ as another condition.
\begin{equation}
\mathrm{AdaGN(h_i,k,f)=h_{s}(k_{s}{\mathrm{GroupNorm}}(f)+k_{b})}
\end{equation}
where $f$ is the normalized feature maps of UNet, and $h_s$ donates the affine projection of $h_i$, while $(k_s,k_b)=\mathrm{MLP}(\varphi(k))$ represents the output of a MLP on sinusoidal encoding of $k$.

Taking in the denoised latent $\tilde{\bf z}^0$,
the trajectory decoder $D_\theta(\tilde{\bf z}^0, h)$ generates future trajectories ${\boldsymbol{\tau}}$ in an autoregressive manner while also incorporating an interaction net to ensure realistic vehicle interactions. The vehicle's acceleration and angle are determined through an iterative process, while the trajectory is updated through a simplistic bicycle model. The state embedding $h_i$ is updated using a recurrent neural network (RNN) as the memory unit. This iterative approach continues until a smooth and coherent path prediction is generated.

\textbf{Training.}
The training of the denoiser $\varepsilon_\theta$ necessitates an auxiliary posterior distribution network $q_\theta(\bf z|{\boldsymbol{\tau}},h)$, which takes future trajectories and current states' context embeddings as inputs to generate training samples $\tilde{\bf z}_i^0$. The posterior network utilizes a network structure similar to that of the context encoder.

We initially train the denoiser and other modules separately, followed by joint training of all modules. To collaboratively optimize the trajectory decoder, context encoder, and posterior net, we employ the modified Evidence Lower Bound (ELBO) loss function commonly utilized in VAEs~\cite{vae}.
\begin{equation}
{\cal L}_{\mathrm{ELBO}}=\mathbb{E}_{q}(\log p({\boldsymbol{\tau}}|{\bf z}))-K L(q({\bf z}|{\boldsymbol{\tau}},h)|p({\bf z}|h))
\end{equation}
Training of the denoising net $\varepsilon_\theta$ is done by optimizing $L_2$ noise loss. Notably, $\tilde{\bf z}^{0}$ is produced from the posterior distribution network.
\begin{equation}
{\cal L}_{\mathrm{noise}}=\sum_{k=1}^{K}\mathbb{E}_{{\tilde{\bf z}}^{0},{\boldsymbol{\epsilon}}^{k}}\Big[||\varepsilon_\theta(\tilde{\bf z}^k,k,{\bf z},h)-{\boldsymbol{\epsilon}}^{k}||_{2}^{2}\Big]
\end{equation}
The final joint training of all modules incorporates collision penalties for traffic flow, in addition to the aforementioned two losses.
\begin{equation}
{\cal L}_{\mathrm{joint}}={\cal L}_{\mathrm{ELBO}}+{\cal L}_{\mathrm{noise}}+{\cal P}_{coll}
\end{equation}

\subsection{Reward Model for Guided Sampling}
% 类条件下guided diffusion公式
Appealing to \cite{diffuser}, reinforcement learning problems can be modeled as a generative process using guided sampling.  In this part, we extend the diffusion model presented in Sec.~\ref{subsec:diff-m} by incorporating a guidance function to generate adversarial vehicle behaviors. Specifically, a binary classifier that identifies adversarial samples is incorporated into the original denoising process transition in Eq.(\ref{equ:transition}).
\begin{equation}
\begin{split}
    &p_{\theta}(\tilde{\bf z}^{k-1}|\tilde{\bf z}^k,O_{1:T};{\boldsymbol{\epsilon}}^{k})\approx N({\boldsymbol{\mu}}^{k-1}+\Sigma g, \Sigma^{k-1})\\
    &with \quad \Sigma^{k-1}=\frac{1-\alpha^{k-1}}{1-\alpha^{k}}\beta^{k}
\end{split}
\end{equation}
where $O_t$ donates a binary indicator of whether the latent code leads to an optimal trajectory at time step $t$ and 
\begin{equation}
    \begin{split}
        g &=\nabla_{\tilde{\bf z}}\log p({\cal O}_{1:T}|\tilde{\bf z})\vert_{\tilde{\bf z}={\boldsymbol{\mu}}}
        =\sum_{t=0}^{T}\nabla{R({\boldsymbol{\mu}},h,{\bf z};{\boldsymbol{\mu}})}\\
        &=\nabla{\mathcal J}({\boldsymbol{\mu}}).
    \end{split}
\end{equation}
where $R(\circ)$ is the reward function associated with the adversarial objective and $\mathcal{J}(\circ)$ computes expectation of accumulated rewards. Thus, adversarial samples can be generated by perturbing the predicted mean during denoising using the gradient of the reward function. Analogous to value-based reinforcement learning, a trainable reward model is formulated to assess the accumulated future rewards ${\mathcal J}({\boldsymbol{\mu}})$.

\textbf{Architecture.}
We adopted the classic DQN network architecture to construct the reward model. Adhering to commonly used notation in reinforcement learning for clarity, we regard vehicle context $(h,{\bf z})$ as states and predicted mean $\boldsymbol{\mu}$ as actions here.
\begin{equation}
\begin{split}
    {\mathcal J}({\boldsymbol{\mu}})&=Q_\theta\bigl(s\,,a\bigr)\\
    with \quad s&=(h,{\bf z}) \ and \ a={\boldsymbol{\mu}}
\end{split}
\end{equation}
where $s$ represents states, $a$ donates actions, and $Q(\circ)$ is the action-value function.
Moreover, for efficient information exchange among vehicles, we integrated the interaction net into the state encoding layer of DQN. Unlike the interaction net employed in trajectory generation, nodes representing target and background vehicles are labeled distinctly.

\textbf{Training.}
We initially train the reward model on data from driving logs by randomly selecting target vehicles within the original scenarios and replaying actions for these target vehicles. Additionally, we generate trajectories of other vehicles by feeding a random $z$ into the trajectory decoder, thereby collecting a series of experiences $(s_t, a_t, r_t, s_{t+1})$ for learning the Q function. Similar to DQN, we use mean squared error here.
\begin{equation}
    {\cal L}_{\mathrm{Q}}=\mathbb{E}_{s_t,a_t}\bigl[||R_{t}\,+\,\gamma\cdot\operatorname*{max}_{a}\,Q_\theta\bigl(s_{t+1}\,,a\bigr)-Q_\theta(s_t,a_t)||_{2}^{2}\bigr]
\end{equation}
where $\gamma$ is the discount factor. The reward $r_t$ at timestep $t$ consists of adversarial rewards and penalties on background collisions.
\begin{equation}
    R_t = {\cal R}_{adv}-{\cal P}_{coll} 
\end{equation}
where ${\cal R}_{adv}$ represents collision rewards incurred by target AVs in generated scenarios, while ${\cal P}_{coll}$ denotes penalties for BVs collisions (excluding collisions involving a BV and an AV).  Collisions encompass both inter-vehicle collisions and environmental collisions (such as leaving drivable areas), employing a differentiable collision detection method based on distance measurement following \cite{trafficsim}.

\subsection{Generating Safety-critical Scenarios}
% \begin{algorithm}[t]
% \textbf{Require} context encoder $E_\theta$, trajectory decoder $D_\theta$, reward model $Q_\theta$, scale $\alpha$, diffusion steps $K$, covariances $\Sigma^k$, actions horizon $l$ to take before re-planning.\\
% \While{not done}
% {Observe and update ($m$, $past$)\;
% $h,{\bf z} = E_\theta(m, past)$\;
% Sampling a random $\tilde{\bf z}^K \sim \mathcal{N}(0,I)$\;
% \ForEach{$k$ in K,.....,1}
% {${\boldsymbol{\mu}}={\boldsymbol{\mu}}_\theta(\tilde{\bf z}^k,k,{\bf z},h$)\;
% ${\boldsymbol{\mu}} \leftarrow {\boldsymbol{\mu}} + \alpha \nabla{Q_\theta({\boldsymbol{\mu}}, h, {\bf z};{\boldsymbol{\mu}})}$\;
% $\tilde{\bf z}^{k-1} \sim \mathcal{N}({\boldsymbol{\mu}}, \Sigma^{k-1})$\; }
% ${\boldsymbol{\tau}} = D_\theta(\tilde{\bf z}^0, h)$\;
% Execute first $l$ actions of trajectory ${\boldsymbol{\tau}}$;
% }
% \caption{Safety-critical Scenario Generation}
% \label{alg:1}
% % \vspace{-0.6cm}
% \end{algorithm}

% The overall pipeline for generating adversarial safety-critical scenarios is illustrated in Alg.1. 
Initially, historical trajectories of all vehicles and their surrounding maps are encoded as contextual embeddings. Conditioned in this context, the diffusion model iteratively denoises random noise while simultaneously evaluating the adversarial reward for each outcome. The denoising results are then refined through gradient guidance of the reward to obtain a latent code capable of generating adversarial background traffic flow after K rounds of denoising. Finally, the trajectory decoder restores the final trajectories from latent space. The interaction between background vehicles and the target vehicle continues until the scenario ends, either through collision or reaching the terminal.

%% file: sec/4_exp.tex
\section{Experiments}
\label{sec:formatting}
We next showcase the capabilities of AdvDiffuser in generating worst scenarios. 
In Sec.~\ref{sec:traffic-sim}, we assess its ability to generate diverse and realistic scenarios. Sec.~\ref{sec:acc_gen} analyzes how AdvDiffuser affects tested planners' security and comfort. Sec.~\ref{sec:generalization} discusses transferability among different tested planners and the effectiveness of the few-shot learning. 

\textbf{Dataset.}
We evaluate our methods on the nuScenes\cite{nuscenes} dataset, which consists of 1000 driving scenes, each spanning 20 seconds duration at 2Hz. Following the split and set guidelines in the nuScenes prediction challenge, we split driving logs into 8-second segments, using past trajectories from a 2-second duration to predict future ones for the following 6 seconds. Scenarios are evaluated within these time windows.

\textbf{Baselines}. 
We compare our model with other state-of-the-art methods for generating adversarial driving scenarios
% 写清楚baseline代码设置
\begin{itemize}
    \item \textit{Replay}: rolls out vehicle trajectories from driving logs that are unresponsive to AV actions.
    
    \item \textit{AdvSim}~\cite{advsim}: employs a surrogate AV to iteratively determine optimal adversarial actions, initialized by the output of SimNet. 
    
    \item \textit{Strive}~\cite{strive}: conducts optimization searches over the latent space of TrafficSim.
    
    \item \textit{Adv-RL}~\cite{ade}: an adversarial BV planner with an actor architecture aligned with SimNet.
\end{itemize}

% metrics写清楚参考哪些论文，缩短内容
\textbf{Metrics.}
We highlight realistic safety-critical scenarios and propose a suite of metrics to assess the quality of generated scenarios, as well as evaluate the performance of tested planners on our adversarial scenarios.

\begin{itemize}
\item \textit{Diversity:} metrics include Final Displacement Diversity (FDD) and Minimum Scenarios Final Distance Error (minSFDE), as per \cite{trafficsim, mixsim}, measuring trajectory diversity.
\item \textit{Distribution realism:} Jensen-Shannon divergence (JSD) \cite{igl2022symphony} quantifies distribution gaps in vehicle velocity, acceleration, and time-to-collision (TTC) histograms.
\item \textit{Common sense:} collision rate (CR) computes the percentage of background vehicles involved in collisions, assessing scenario realism.
\item \textit{Adversarial:} metrics evaluate collision severity based on targeted vehicle collision rates and velocities, as well as assessing maneuver comfort via acceleration and jerk.
\end{itemize}

\textbf{Implementation Details.}
AdvDiffuser is implemented using PyTorch and trained on 4 GeForce RTX 3090 GPUs. The diffusion model is trained for 200 epochs using the Adam optimizer with a learning rate of $5*10^{-4}$. For the reward model, we employ a learning rate of $10^{-3}$ and conduct 100 epochs for pretraining. The batch size is set to be 8.

\begin{table*}
 \renewcommand{\arraystretch}{1.0}
    \centering
    \caption{Evaluation of real traffic simulation. Our diffusion model without a guide generates diverse and realistic traffic flow. The observed elevated collision rate with the environment can be attributed to the inclusion of roadside parked vehicles in the original dataset. }
\resizebox{\linewidth}{!}{
    \begin{tabular}{l|ccccccc}
    \toprule
         \multirow{2}{*}{Algorithms}  
             &\multicolumn{2}{c}{\textbf{Diversity}} &\multicolumn{3}{c}{\textbf{Distribution Realism (JSD)}} &\multicolumn{2}{c}{\textbf{Common Sense}}\\
             &minSFDE~($m$)  &FDD~($m$) &{Vel}~($e^{-2}$)  &{Acc}~($e^{-2}$)  &{TTC}~($e^{-2}$) &Env Coll~($\%$) &Veh Coll~($\%$)\\
            \cmidrule(r){1-1}\cmidrule(r){2-3}\cmidrule(r){4-6}\cmidrule(r){7-8}
        
         \textit{AdvSim}&5.49  &3.12  &0.89  &2.21 &0.95 &13.06 &0.87\\
         \textit{Strive} &2.70 &14.48 &0.96 &2.33 &1.85 &12.02 &0.16\\
         \textit{Ours~(traj2traj)} &931  &\textbf{3549}  &21.27  &14.82 &16.63 &22.82 &0.69\\
         Ours &\textbf{2.21}  &14.66  &\textbf{0.81}  &\textbf{1.68} &\textbf{0.57} &\textbf{11.89} &\textbf{0.02}\\
    \bottomrule
    \end{tabular}
}
\vspace{-0.3cm}
    \label{tab:real-sim}
\end{table*}

\subsection{Simulating Real Traffic}
\label{sec:traffic-sim}
% \begin{figure*}[t]
% \includegraphics[width=\linewidth]{figs/regular-sample.png}
%    \caption{Simulated accident-prone scenarios sampled from AdvDiffuser. }
% \label{fig:regular-sample}
% \end{figure*}
To evaluate the performance of AdvDiffuser in simulating real traffic, we assess the diversity and plausibility of generated trajectories. Our model is compared with the generative baselines, \textit{AdvSim} and \textit{Strive} for conventional traffic flow generation. To demonstrate the feasibility of diffusion in the latent space, we also compare its performance with that of directly applying diffusion on trajectories, marked as "traj2traj".
For sampling models, ten samples are generated for each initial scene. To ensure comparability, all models utilize an identical GNN-based context encoder. 

Tab.~\ref{tab:real-sim} shows the quantitative results.
In contrast to the individual control exerted by \textit{AdvSim}, the collective control of multiple agents produces motion distributions that closely resemble real-world patterns, characterized by reduced JSD values and minimal inter-vehicle collision rates. Notably, diffusion on trajectories shows higher FDD values indicating greater diversity of generated scenarios but also exhibits the largest MinSFDE value suggesting significant deviation from real-world scenarios. 
This observation suggests that the direct diffusion along trajectories may pose challenges and hinder the coverage of original real-world scenarios.
Our method integrates the reverse diffusion process over the latent space, yielding superior performance with a slight improvement over \textit{Strive}. This improvement can be credited to the diffusion process, gathering samples that closely mimic the real-world posterior distribution and enriching diversity.

\subsection{Generating Accident-Prone Scenarios}
We show that AdvDiffuser creates challenging scenarios leading to collisions and discomfort for the tested planner, along with more realistic background traffic flows. 
Since rule-based planners remain prevalent in practical AV systems, we use a simple lane-graph-based planner~\cite{montemerlo2008junior} as the attacked planner. \textit{AdvSim} and \textit{Strive} are both optimized for 20 rounds and use the rule-based planner as their surrogate model.
\textit{Adv-RL} and the reward model of AdvDiffser adopt the same network structure.

\begin{table*}
 \renewcommand{\arraystretch}{1.0}
\centering
% \vspace{-0.6cm}
\caption{Evaluation of generated safety-critical scenarios compared with existing algorithms. The attacked planner is a rule-based one. AdvDiffuser achieves comparable adversarial performance while exhibiting advantages in terms of scenario rationality and efficiency, demonstrating a more comprehensive and balanced performance.}
\resizebox{\linewidth}{!}{
    \begin{tabular}{l|cccccccc}
    \toprule
         \multirow{2}{*}{Algorithms} 
            &\multicolumn{2}{c}{\textbf{Collision}}  &\multicolumn{2}{c}{\textbf{AV Comfortable}}  &\multicolumn{3}{c}{\textbf{BV Plausibility}} &\multicolumn{1}{c}{\textbf{Real Time}}\\
            &CR~($\%$) &{Coll Vel~($m/s$)}  &Acc~($m/s^2$)  &Jerk~($m/s^3$)  &{JSD~($e^{-2}$)}  &Env Coll~($\%$) &Veh Coll~($\%$) &$>10Hz$ \\
            \cmidrule(r){1-1}\cmidrule(r){2-3}\cmidrule(r){4-5}\cmidrule(r){6-8}\cmidrule(r){9-9}
            % \cmidrule(r){10}
         % \hline
         \textit{Replay} &8.45  &7.73  &4.13  &12.98  &0.14  &16.86 &0.0 &- \\
         \textit{AdvSim} &\textbf{18.54}  &4.72 &\textbf{4.28}  &\textbf{13.66}  &3.48  &16.83 &0.94 &\XSolidBrush\\
         \textit{Strive} &10.33  &5.88  &4.27  &13.72  &2.64  &16.04 &0.43 &\XSolidBrush \\
         \textit{Adv-RL} &10.56  &5.46  &3.71  &10.91  &7.48  &21.71 &5.21 &\Checkmark \\
         Ours &11.03&\textbf{7.26}&3.89&11.97&\textbf{2.49}&\textbf{14.91}&\textbf{0.27}&\Checkmark\\
    \bottomrule
    \end{tabular}
}
    %advdiffuser产生了一个相对均衡的结果，效率、碰撞和合理性
    \label{tab:my_label}
    % \vspace{-0.6cm}
\end{table*}

Tab.~\ref{tab:my_label} presents the results of the quantitative evaluation. 
Compared to real-world driving scenarios, \textit{Replay} generates slightly more challenging situations, lacking responsiveness that actively avoids AV encounters. \textit{Replay} can serve as a bottom line when evaluating the adversarial performance.
Search-based methods, \textit{AdvSim} and \textit{Strive}, have been observed to increase collision rates and worsen accident severity in AVs, causing discomfort from sudden acceleration and jerkiness. However, the deviation in distribution and collision rates of BVs worsens compared to free-flow traffic simulations, attributed to their inclination towards adversarial aspects rather than rational behavior during the optimization process. Notably, \textit{AdvSim} outperforms \textit{Strive} in adversarial scenarios but has worse trajectory rationality, possibly due to its direct search on actions, which is more flexible. While \textit{Strive}'s exploration of a condensed latent space enhances the plausibility.
It is worth noting that both methods require significantly more time than other methods due to the planner-on-loop searching process, preventing real-time testing.
By incorporating a guided sampling mechanism, AdvDiffuser effectively trains collective adversarial attacks on a comparative scale of adversarial parameters to \textit{Adv-RL}. Nevertheless, AdvDiffuser outperforms the individual control one, \textit{Adv-RL}, across nearly all metrics.
Overall, AdvDiffuser demonstrates comparative adversarial performance, with collision rates surpassed only by \textit{AdvSim}, while outperforming in accident severity, scenario plausibility, and efficiency compared to alternatives, highlighting its well-rounded and balanced performance for superior outcomes.
These observations are quantitatively supported by Fig.~\ref{fig:illustration}. 

% \begin{figure*}[htbp]
% \centering
% \includegraphics[width=0.95\linewidth]{figs/adv-samples.png}
%    \caption{Accident-prone scenarios sampled from AdvDiffuser, including rear-end collisions, cutting in, lane changes, and pulling into oncoming traffic. The targeted vehicles are marked with green squares, and attackers are represented by orange squares. Original scenario trajectories are shown with white lines, while AdvDiffuser-generated modified trajectories are depicted with linked colorful circles.}
% \label{fig:adv-sample}
% % \vspace{-0.6cm}
% \end{figure*}

\subsection{Transferability Analysis}
\label{sec:generalization}
\begin{figure}[t]
\centering
\includegraphics[width=0.8\linewidth]{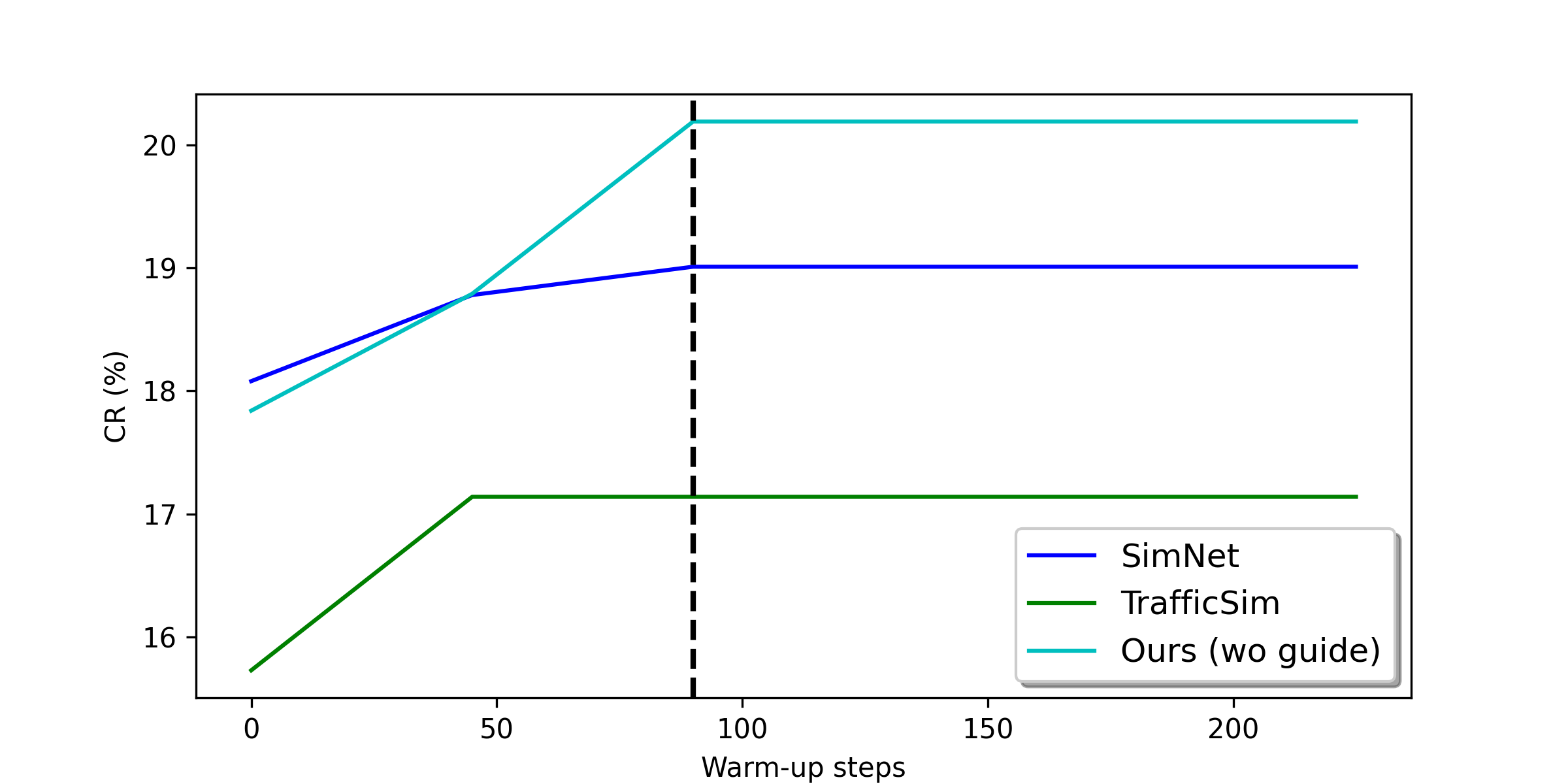}
   \caption{Evaluation of few-shot learning. AdvDiffuser exhibits increased collision rates when encountering novel targeted planners following a few initial interactive steps.}
\label{fig:transfer}
\vspace{-0.5cm}
\end{figure}

We assess the transferability of adversarial methods across different planners, as shown in Tab.~\ref{tab:transfer}. In addition to the rule-based planner, we evaluate various traffic simulation planners as discussed in Sec.~\ref{sec:traffic-sim}. Initially, we examine the performance of AdvDiffuser in a few-shot learning. As depicted in Fig.~\ref{fig:transfer}, even with a limited number of warm-up steps, we consistently improve adversarial effectiveness, resulting in increased collision rates for all new targets within 90 steps.

\begin{table*}
\centering
\caption{Transferability of safety-critical scenario generation methods across diverse target planners. AdvDiffuser maintains stable evaluation outcomes while ensuring a reasonable background traffic flow.}
\renewcommand{\arraystretch}{1.0}
    \resizebox{\linewidth}{!}{
    \begin{tabular}{lc|ccc|ccc|ccc}
    \toprule
    \multicolumn{2}{c|}{\multirow{2}{*}{Algorithms}} 
        &\multicolumn{3}{c|}{\textbf{Rule-based}} &\multicolumn{3}{c|}{\textbf{SimNet}} &\multicolumn{3}{c}{\textbf{TrafficSim}} \\
         \multicolumn{2}{c|}{} &\textit{SimNet}  &\textit{TrafficSim}  &Ours~(w/o guide)  &\textit{Rule-based}  &\textit{TrafficSim} &Ours(w/o guide)  &\textit{Rule-based}  &\textit{SimNet} &Ours~(w/o guide)  \\
         
    \cmidrule(r){1-2}\cmidrule(r){3-5}\cmidrule(r){6-8}\cmidrule(r){9-11}
        
    \multirow{2}{*}{\textit{AdvSim}}  &AV CR~($\%$) &\underline{20.89}&20.89&20.66 
    &$\mathbf{16.20~(13\%\downarrow)}$ &31.69&23.00 
    &$\mathbf{15.96~(14\%\downarrow)}$&\underline{33.33}&22.54\\ 
    &BV CR~($\%$) 
    &17.77&17.77&17.77
    &17.73&16.16&17.73
    &17.73&16.82&17.93\\ 

    \cmidrule(r){1-2}
    \multirow{2}{*}{\textit{Strive}}  &AV CR~($\%$) &18.54&16.43&\underline{18.77} 
    &$\mathbf{9.56~(7\%\downarrow)}$ &15.49&\underline{18.08} 
    &$\mathbf{10.03~(3\%\downarrow)}$ &19.48&\underline{19.95}\\ 
    &BV CR~($\%$) 
    &16.35&16.39&16.44 
    &16.46&16.51&16.43 
    &16.45&16.54&16.37 \\ 
    
    \cmidrule(r){1-2}           
    \multirow{2}{*}{\textit{Adv-RL}} &AV CR~($\%$)
    &\underline{22.30}&20.66&21.13 
    &$\mathbf{10.74~(4\%\uparrow)}$&23.24&23.71 
    &$\mathbf{9.62~(9\%\downarrow)}$&\underline{27.93}&26.06\\ 
    &BV CR~($\%$) 
    &26.91&26.9&26.92 
    &29.31&29.31&29.31 
    &30.75&30.7&30.75\\
    
    \cmidrule(r){1-2}  
    \multirow{2}{*}{Ours}  &AV CR~($\%$) 
    &19.01&17.14&\underline{20.19}
    &$\mathbf{11.03~(\approx)}$ &15.49&\underline{16.67} 
    &$\mathbf{11.50~(4\%\uparrow)}$ &17.37&\underline{17.84} \\ 
    &BV CR~($\%$) 
    &14.78 &14.79 &14.73 
    &14.63&14.81&15.25 
    &14.71&14.98&14.94\\ 
    \bottomrule     
    \end{tabular}
}
    \label{tab:transfer}
\vspace{-0.3cm}
\end{table*}
\label{sec:acc_gen}

Moreover, we thoroughly assess the transferability of AdvDiffuser and other SoTAs. Notably, \textit{Adv-RL} and \textit{SimNet} consistently demonstrate superior attack outcomes against \textit{AdvSim}, while \textit{Strive} and AdvDiffuser show optimal attacks against our fundamental diffusion model.
The strong correlation between an attacker's optimal outcomes and a structurally similar planner supports the intuition that better accuracy in predicting the targeted vehicle leads to more successful attacks.
We highlight the significance of columns related to \textit{rule-based}, as \textit{rule-based} shows a low structural correlation with other neural planners and demonstrated the highest resilience against attacks from all these neural adversarial models with the lowest collision rate.
This unique attribute makes it a relatively ideal test target for assessing transferability across different source planners.
While other adversarial techniques exhibit a slight degradation in performance compared to results trained on \textit{rule-based}. AdvDiffuser stands out by maintaining consistent and stable evaluation outcomes while ensuring a reasonable background traffic flow.

%% file: sec/5_dis.tex
\section{Discussion}

The absence of a universally acknowledged benchmark for safety-critical driving scenarios poses significant concerns regarding the practical utility of simulation-generated adversarial scenarios on real-world autonomous driving. The validity of using such scenarios as evaluation criteria warrants further investigation, particularly regarding their relevance to actual driving conditions. The pivotal question revolves around discerning the extent to which unrealistic or highly improbable hazardous situations contribute to the enhancement of autonomous driving system safety. Redirecting attention towards assessing the likelihood of a simulated scenario manifesting in real-world settings may offer more pragmatic insights for understanding and refining the efficacy of autonomous driving systems.